# When Explanations Cannot Be Read: Measuring and Correcting SHAP and LIME Rendering for Right-to-Left Languages

**Rameesha Zia**
ORCID: 0009-0009-6717-2564
*School of Computing Sciences, Pak-Austria Fachhochschule, Institute of Applied Sciences and Technology, Haripur 22620, Pakistan*
rameeshaazia@gmail.com

**Muhammad Shahid Iqbal Malik***
ORCID: 0000-0001-8396-3344
*School of Computing Sciences, Pak-Austria Fachhochschule, Institute of Applied Sciences and Technology, Haripur 22620, Pakistan*
shahid.malik@paf-iast.edu.pk

## Abstract

Post hoc explanation methods such as SHAP and LIME are widely used to interpret text classifiers, but their visualizations are mainly designed for left-to-right languages. When applied to right-to-left (RTL) languages such as Urdu, Arabic, Persian, and Hebrew, the attribution values remain mathematically valid, while their visual presentation fails. Tokens appear out of sequence, connected letterforms break apart, and plot layouts do not follow the natural reading direction. This study addresses this gap as a visualization problem rather than a limitation of the explanation methods themselves. We present SHAP-RTL, a rendering layer that corrects reading direction and script shaping in SHAP and LIME visualizations, with per-language font selection, while preserving the original attribution values, feature ordering, and model outputs. The approach is evaluated on Urdu, Arabic, Hebrew, and Persian hate and offensive-language datasets using TF-IDF and logistic regression classifiers. Rendering correctness is measured by an OCR round trip over 200 feature words per language. Default rendering yields character error rates of 0.820 to 0.979, meaning the label no longer carries its token; the common reshape-and-reorder workaround fails for Urdu at 0.998, worse than no correction; and the Matplotlib 3.11.0 text rewrite inverts that workaround, while SHAP-RTL remains correct under both versions. The framework also verbalizes the same attributions as short contextual explanations in the reader's language, constrained to the identified features. Evaluation in this paper concerns rendering correctness; assessment of the generated explanations is left to future work. The study highlights the importance of language-aware visualization in making post hoc explainability more accessible across different writing systems.

## Nomenclature

| | | | |
|---|---|---|---|
| ML | Machine Learning | LTR | Left to Right |
| DL | Deep Learning | RTL | Right to Left |
| SM | Social Media | NLP | Natural Language Processing |
| XAI | Explainable Artificial Intelligence | UBA | Unicode Bidirectional Algorithm |
| SHAP | SHapley Additive exPlanations | LIME | Local Interpretable Model-agnostic Explanations |
| CTL | Complex Text Layout | LLM | Large Language Model |
| HS | Hate Speech | OL | Offensive Language |
| LR | Logistic Regression | TF-IDF | Term Frequency-Inverse Document Frequency |
| OCR | Optical Character Recognition | CER | Character Error Rate |
| GSUB | Glyph Substitution table | GPOS | Glyph Positioning table |
| | | | |

## 1. Introduction

Machine learning systems, and DL models in particular, have now been widely used for text classification tasks such as sentiment analysis, content moderation, and SM monitoring. These models perform well, but their decisions are often difficult to interpret and understand [1]. Post hoc explainability methods were introduced to address this problem without changing the trained model. Rather than building interpretability into the architecture itself, these methods fix the trained model, and construct an explanation layer around its output [2]. Among these methods, SHAP [3] and LIME [4] have become the two most widely adopted techniques. Both can show how individual input features, such as tokens in text, contribute to a prediction.

Although SHAP and LIME are designed to interpret model behaviour independently of language, their visual output assumes left-to-right reading order. Applied to RTL languages such as Urdu, Arabic, Persian and Hebrew, tokens are drawn out of sequence, cursive letterforms are broken apart, and the plot layout does not follow the direction in which the text is read [5]. The underlying SHAP values and LIME coefficients remain mathematically valid; the failure is in how they are displayed.

This is a human readability problem rather than a computational one, and it has received little attention in the explainability literature, which has focused almost entirely on English and other LTR languages [6]. Until recently, SHAP and LIME visualization libraries provided no built-in handling of reading direction or the shaping requirements of cursive scripts. As Section 3.4 documents, the underlying plotting library changed this in mid-2026, but the change is recent, is not yet reflected in deployed environments, and does not address per-language font selection or layout geometry. It also silently invalidates the workaround the community had adopted in the interim. Mixed-direction text adds a further complication, since RTL social media posts routinely contain Latin-script terms, hashtags and numerals whose placement is resolved by the bidirectional algorithm; the package handles these, though the present evaluation covers single-script labels.

To overcome this, we propose a rendering layer for SHAP and LIME that applies complex text layout and per-language font selection at the label level while leaving attribution values, feature ordering and model outputs unchanged. We evaluate it quantitatively on Urdu, Arabic, Hebrew and Persian hate and offensive-language datasets, measuring rendering correctness directly rather than by inspection. Urdu serves as the primary case study, given its combination of Nastaliq shaping and its comparatively limited representation in existing NLP tooling.

The contributions of this work are as follows.

- First, we characterize the rendering failures that occur when SHAP and LIME visualizations are applied to right-to-left languages, and categorize them into three mechanisms: incorrect text direction, shaping failure, and uneven layout. We examine these issues across four languages and eight plot families.
- Second, we introduce a quantitative measure of rendering correctness for attribution plots. Rendering correctness is assessed by an OCR round trip over 200 feature words per language, which to our knowledge has not previously been applied to explanation visualizations.
- Third, we show that the reshape-and-reorder workaround in common community use fails for Urdu at a character error rate of 0.998, worse than applying no correction at all, and identify the cause: it substitutes Unicode presentation forms that contemporary Nastaliq fonts do not contain.

- Fourth, we show that the Matplotlib 3.11.0 text rewrite repairs default rendering for Arabic and Persian while silently inverting that same workaround in all four languages, so code that adopted it degrades on upgrade with no error raised.
- Fifth, we provide an installable rendering layer that performs complex text layout and per-language font selection for SHAP and LIME plots, returns identical error rates under both library versions, and preserves attribution values, feature ordering and model outputs. The package additionally verbalizes attributions as contextual explanations in the reader's language; this component is provided but not evaluated here.

The remainder of the paper is organized as follows. Section 2 reviews post hoc explainability methods with a focus on SHAP and LIME, and traces where their assumptions about text direction originate. Section 3 characterizes the misalignment problem across the four languages under study, decomposes it into three failure mechanisms, and establishes the library versions to which each applies. Section 4 states the research questions. Section 5 describes the proposed rendering layer and the language-model explanation stage. Section 6 covers the datasets, classifiers and evaluation protocol. Section 7 reports rendering correctness under two library versions and presents the paired visual comparisons. Section 8 concludes.

## 2. Post hoc XAI methods

Explainability in ML divides broadly along two lines: models that are interpretable by design, and models whose decisions are explained only after the fact. Post hoc methods belong to the second category. Instead of making the model itself easy to understand, they take the trained model as given and produce a separate explanation for its predictions [6]. A prediction is generated first; an explanation describing the reason for that prediction is constructed afterwards. This separation is what makes post hoc explanation attractive for DL and transformer-based architectures, since it allows researchers to keep high-performing black-box models while still recovering some account of their internal reasoning.

Post hoc methods are typically grouped into families that differ in how they arrive at an explanation and in what they treat as the unit of analysis, as summarized below; **Fig. 1** shows the relative adoption of ten widely used methods [7]. Perturbation-based approaches such as LIME and SHAP add noise to the input, observe the resulting change in output, and estimate feature importance from that relationship [8]. Gradient-based methods, including Saliency Maps, Integrated Gradients and Grad-CAM, propagate gradients backwards to track the sensitivity of the output to each input dimension [9]. Attention-based methods read the attention weights produced by a model as an indication of feature importance; these are especially common in NLP, where attention mechanisms are central to transformer architectures [6]. Surrogate-based methods approximate the black box with a simpler interpretable model, either locally around a single prediction or globally across the input space [10]. Counterfactual methods take a different route entirely, explaining a prediction by showing the minimal input change that would have produced a different outcome [11].

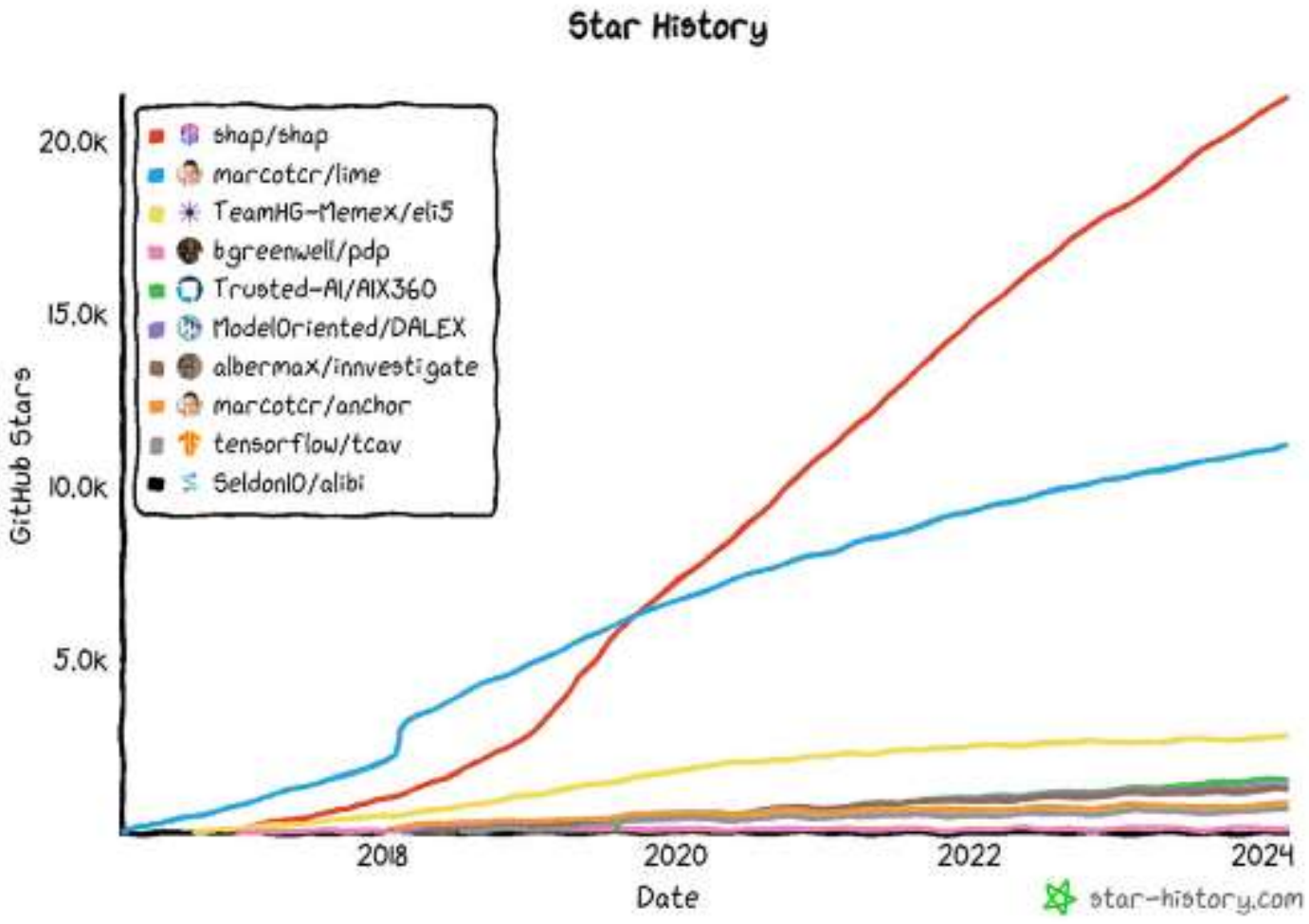

**Fig. 1**: GitHub star history for ten common XAI methods. Reproduced from Salih et al. [7] under CC BY 4.0.

## 2.1 SHAP

SHAP was introduced by Lundberg and Lee [3] as a way of unifying several existing explanation methods, including LIME, under one theoretical roof. The central idea comes from cooperative game theory: The Shapley value, a classical method for fairly dividing a payoff among players according to how much each contributes across every possible order of entry into the game. Lundberg and Lee recast a model's prediction in these terms, treating each feature as a player and asking how much that feature contributes to the outcome.

The Shapley value of feature *i* for model *f* and feature set *N* can be formally defined as:

$$\varphi_i = \sum_{S \subseteq N\{i\}} \frac{|S|!\ (|N|! - |S|! - 1)!}{|N|!} [f(S \cup \{i\}) - f(S)] \quad (1)$$

where $\varphi_i$ is the attribution assigned to feature *i*, *N* is the full set of features, *S* ranges over all subsets of *N* that exclude *i*, and *f*(*S*) is the model output when only the features in *S* are present. Equation (1) is the average of the marginal effect of adding feature *i* over all possible subsets of the other features. When computed in this manner, the values add up exactly to the difference between the actual prediction from the model and its baseline prediction, a property known as local accuracy. The problem with computing this exactly, however, is that it quickly becomes intractable for more than a few tokens of the input. In practice, therefore, researchers use approximations: KernelSHAP performs a weighted regression of the values over sampled feature coalitions, TreeSHAP is efficient and exact for tree-based models [12], and DeepSHAP extends the concept to neural networks by back-propagating attributions through the network.

In text classification, SHAP looks at each token in a sentence and shows how much that token affects the model's final decision. It does this by masking or removing some tokens and checking how the model's output changes. The result is a value for each token, which can then be shown in plots such as heat maps and force plots.

## 2.2 LIME

LIME was proposed by Ribeiro, Singh and Guestrin [4] out of a concern that most classifiers offer no transparency into their decisions. Their solution was not to explain a model as a whole but to explain one prediction at a time. LIME takes a single instance, generates many perturbed variants of it, weights those variants by their distance from the original, and fits a simple linear model over the resulting local neighborhood. The coefficients of that small linear model constitute the explanation: which features exerted influence, in which direction, and at what magnitude.

Applied to text, this usually means deleting words from a sentence and observing how the classifier's confidence shifts. Words whose removal moves the prediction most receive the largest weight in the local surrogate, and those are the words LIME highlights. Its main practical advantage is model-agnosticism combined with speed relative to exact SHAP, which keeps it popular despite documented instability across repeated runs [7].

## 2.3 Locating the directional assumption

The assumption of left-to-right reading order does not reside in SHAP or LIME. Both operate on feature indices and numerical attributions, and neither has any representation of script or reading direction. What SHAP and LIME emit is a set of label strings and values that are passed to a general-purpose plotting library, which draws the text through its own font and layout path. Whether a label is shaped and ordered correctly is therefore determined by that path, not by the attribution method.

This is why the defect appears identically in SHAP and LIME, why it cannot be corrected by changing how attributions are computed, and why a change to the plotting library's text handling can resolve it without either method being modified. Section 3 examines that path directly, separating the resulting failures into three mechanisms and establishing the library versions in which each occurs.

## 3. Problem Analysis

Most post hoc explainability tools, including the visualization frameworks used by SHAP and LIME, were developed with English and other LTR languages in mind. Their visualizations therefore follow LTR conventions by default, affecting token order, the presentation of attribution bars, and the arrangement of force plots. RTL languages were simply not part of the design space.

The misalignment occurs as soon as these techniques are applied to any RTL language. These languages are not merely read in the opposite direction. Both SHAP and LIME assume a single uniform flow from first token to last, left to right, and render explanations on that basis regardless of the language being processed. This behavior is documented in open reports against the plotting library itself, covering reversed and disconnected Arabic and Hebrew across titles, axis labels, legends and annotations [13].

In practice this means that, under the plotting configurations in common use at the time of writing, an explanation generated for RTL text (Urdu, Arabic, Hebrew, or Persian) is frequently rendered in a form that does not correspond to the sentence it explains. The attribution bars are in the opposite direction compared to the reading order. For a native speaker of any of these four languages, the resulting plot is not a faithful representation of the explanation but a visual artefact that reads backwards, breaks words apart, or otherwise violates the conventions of the script. The problem is one of human readability rather than computation. The Shapley values and local surrogate coefficients underlying SHAP and LIME are mathematically valid for any script to which they are applied. What fails is the presentation of those values.

This rendering failure has gone almost unremarked in the explainability literature, which has focused overwhelmingly on English and other left-to-right languages. Hundreds of millions of people use right-to-left languages, yet XAI research has paid them very little attention, and current XAI tooling still lacks proper support. Nor is the gap confined to the explainability stack. The problem we describe in SHAP and LIME is one instance of a systemic gap that runs through the Python visualization ecosystem, which is part of why patching individual call sites has not resolved it.

### 3.1 Three failure mechanisms

After examining the default SHAP and LIME plots for the four languages, we found three main types of RTL rendering problems. These problems are different from each other, so each one needs a different type of fix.

1. The first problem is directional inversion. RTL text is stored in one order but needs to be displayed from right to left. When the plotting system does not handle this correctly, the glyphs are drawn in logical rather than visual order.
2. The second problem is shaping failure. This is especially important for Arabic-based scripts such as Arabic, Urdu, and Persian. In these scripts, letters can change their shape depending on their position in a word and can connect with nearby letters. If this shaping is not handled correctly, the letters may appear separated or broken, producing isolated forms that do not correspond to the written word. Urdu can be even more challenging because it commonly uses the Nastaliq writing style, which has more complex letter connections and positioning. Hebrew is less affected by this problem because its letters are not connected in the same way as Arabic-based scripts.
3. The third problem is layout asymmetry. Even when the text is displayed correctly, the overall plot may still follow a left-to-right layout. For example, in a horizontal bar plot, labels are normally placed on the left and the bars extend to the right. This places the label on the side an RTL reader encounters last. Force plots can have a similar issue because their visual flow is also based on a left-to-right layout.

**Table 1** summarizes these three problems across the plot types examined in this study. ● indicates the mechanism is present in the default output; ○ indicates it is absent or negligible. The rendering layer described in Section 5 addresses directional inversion and shaping failure; layout asymmetry is not resolved by the present implementation and is discussed in Section 8. **Fig. 2** and panels (a) and (c) of **Figs. 3–10** show examples of the default output for the four languages before applying our RTL rendering approach.

### 3.2 Observed misalignment across the four languages

This section discusses the results of SHAP and LIME, when applied to right-to-left text. All the plots in this section are produced by the standard libraries without making any changes to the original output. The results showed that the problem is not limited to one language or one tool. Instead, similar rendering problems appear across different RTL languages and both explainability methods. Urdu is included in every SHAP comparison because it is the hardest of the four languages to render correctly (Section 7.1, Table 5).

**Table 1**: Rendering failure mechanisms observed by plot family and language group. ● indicates the mechanism is present in the default output; ○ indicates it is absent or negligible.

| Plot family | Level | Directional inversion | Shaping failure (Arabic script) | Shaping failure (Hebrew) | Layout asymmetry |
|---|---|---|---|---|---|
| Summary | Global | ● | ● | ○ | ● |
| Bar | Global | ● | ● | ○ | ● |
| Beeswarm | Global | ● | ● | ○ | ● |
| Heatmap | Global | ● | ● | ○ | ● |
| Waterfall | Local | ● | ● | ○ | ● |
| Force | Local | ● | ● | ○ | ● |
| Decision | Both | ● | ● | ○ | ● |
| LIME bar | Local | ● | ● | ○ | ● |

SHAP summary plots showing the misaligned keywords for Urdu (a) and Arabic (c) are presented in **Fig. 3**. Both languages use the Arabic script, but the problems are more noticeable in Urdu. The Arabic labels in (c) contain recognizable but unjoined letters in reversed order. In (a), the Urdu labels are harder to read because Urdu commonly uses the Nastaliq writing style. The letters are connected and placed along a sloping line, so incorrect rendering can make the whole word difficult to recognize.

A comparison of Urdu and Hebrew bar plots shows that RTL rendering problems are not the same for every script, as presented in **Fig. 4**. Hebrew letters are generally written separately, so the main problem in (c) is the incorrect reading direction. In (a), Urdu shows both problems: the text appears in the wrong order, and the letter shapes are not rendered correctly. This shows that multiple RTL rendering problems can occur at the same time.

A comparison of the SHAP waterfall plots for Urdu and Persian shows rendering problems in both languages, as presented in **Fig. 5 (a), and (c)**. Both languages use an Arabic-based script, but Urdu commonly uses the Nastaliq style, while Persian is often displayed in a Naskh-style font. Despite these differences, both languages show similar rendering problems. This suggests that the issue is not limited to one language or one writing style. In the waterfall plot, labels are placed close to numbers and other visual elements. When the text is difficult to read, it becomes harder to identify which value belongs to which feature.

The SHAP force plot for Urdu in **Fig. 6** shows another type of rendering problem. The issue is more noticeable because the feature names are placed inside the colored parts of the plot, where there is limited space. As a result, broken or incorrectly ordered text becomes more difficult to read. The force plot also follows a clear visual direction from the starting value to the final prediction, which does not naturally match the right-to-left reading direction.

The LIME results for all four languages are discussed here, with Urdu and Arabic in **Fig. 9(a)** and **Fig. 10(a)**, and Persian and Hebrew in **Fig. 2**. The same types of problems that were seen in the SHAP plots also appear here. Urdu and Arabic have problems with the text order and broken letter shapes. These issues are more visible in Urdu. Persian shows similar problems, while Hebrew mainly has an incorrect text direction without broken letters. The same failures appear in both SHAP and LIME despite their different attribution mechanisms, consistent with Section 2.3: the defect lies in the shared Matplotlib text path.

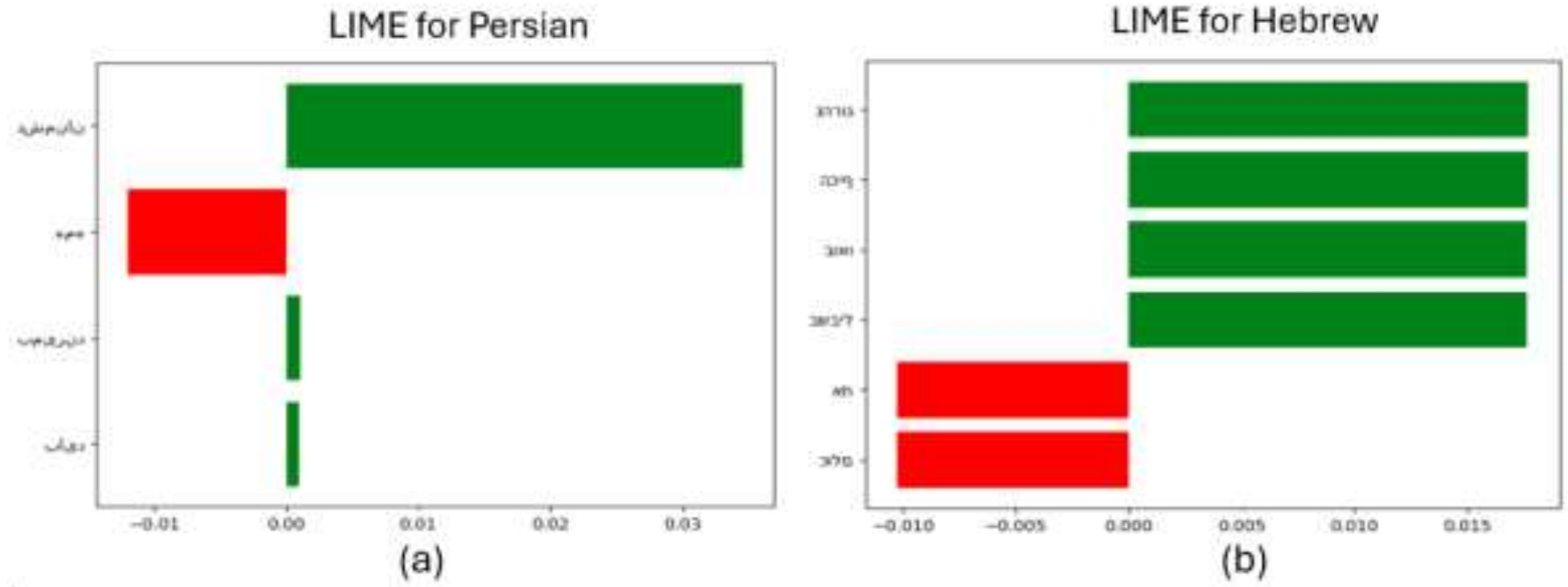

**Fig. 2**: Default LIME Bar plot representing misaligned keywords in (a) Persian and (b) Hebrew.

### 3.3 Scale of the affected population

The use of Right-to-left languages goes beyond the four languages studied here. **Table 2** shows this range across thirteen languages, including their regions, number of speakers and presence on SM. Even so, a small number of these languages account for most right-to-left speakers worldwide. Arabic alone has an estimated 370 to 450 million speakers, and together with Urdu and Persian, these three represent well over half a billion speakers combined [14-17], which is why the paper centers on them rather than spreading attention across all thirteen.

**Table 2:** RTL language demographics and social media usage across the thirteen languages surveyed.

| Language | Primary countries / regions | Speakers (millions) | Social media platforms with active usage |
|---|---|---|---|
| Arabic | 22+ countries (Saudi Arabia, Egypt, UAE, Iraq, etc.) | 370–450 | YouTube, WhatsApp, Facebook, X (Twitter), TikTok, Snapchat |
| Urdu | Pakistan, India | 230 | WhatsApp, YouTube, Facebook, TikTok, Instagram |
| Persian (Farsi) | Iran, Afghanistan (as Dari), Tajikistan | 110 | Instagram, Telegram, WhatsApp, X (Twitter), YouTube |
| Pashto | Afghanistan, Pakistan | 50–60 | Facebook, WhatsApp, YouTube, TikTok |
| Kurdish | Iraq, Iran, Turkey, Syria | 30–40 | Facebook, Instagram, WhatsApp, Telegram |
| Sindhi | Pakistan, India | 30–35 | Facebook, WhatsApp, YouTube |
| Azeri (Arabic script) | Iran (Azerbaijan region) | 15–20 | Instagram, Telegram, WhatsApp |
| Uyghur | China (Xinjiang), Kazakhstan | 12 | WeChat, Douyin, Telegram |
| Balochi | Pakistan, Iran, Afghanistan | 10 | WhatsApp, Facebook, YouTube |
| Hebrew | Israel | 9 | Facebook, WhatsApp, Instagram, LinkedIn, TikTok |
| Kashmiri | India (Jammu & Kashmir), Pakistan | 7 | Facebook, WhatsApp, X (Twitter) |
| Aramaic (Syriac) | Iraq, Syria, Turkey, Iran | 1–2 | Facebook, WhatsApp (community groups) |
| Divehi | Maldives | 0.35 | Facebook, Viber, Instagram, X (Twitter) |

Information about the RTL languages, their speakers, population and SM platforms in which these languages are used are referred: [14-17]

Hebrew is included as a fourth language not for its speaker count, which is comparatively small at roughly nine million, but because it offers established NLP resources and an annotated dataset [18], and because its orthographic properties differ from Arabic, Urdu and Persian despite the shared reading direction. Hebrew is written with non-connecting letterforms, which isolates the directional and layout mechanisms from the shaping mechanism and gives the case study a genuine control condition rather than three structurally similar languages. The remaining nine languages in **Table 2** illustrate the wider scale of right-to-left usage and motivate why the misalignment problem reaches beyond the four examined directly.

### 3.4 Version scope of the reported defect

The rendering behavior reported in this section was observed under Matplotlib 3.10.8 and earlier, the version range in widespread deployment at the time this study was conducted, including the hosted notebook environments in which the default outputs shown in Fig. 2 and the default panels of Figs. 3–10 were produced. Matplotlib 3.11.0, released in June 2026, rebuilt text and font processing on libraqm, HarfBuzz and SheenBidi [19], which addresses directional ordering and contextual joining for text passed directly to the plotting layer. Two problems persist independently of version. First, the plotting library performs no per-language font selection, so Urdu continues to render in Naskh letter forms unless a Nastaliq font is supplied explicitly, which neither the plotting library nor the explanation libraries do. Second, the workaround most widely recommended in practice, contextual reshaping followed by bidirectional reordering, becomes actively harmful under 3.11.0 because the Unicode presentation forms it substitutes conflict with the new shaping path. Section 7.1 quantifies both the original defect and this regression, and reports every measurement with the version that produced it.

## 4. Problem statement and research questions

The problem addressed in this work has two main parts. We discuss them separately because they require different solutions.

The first concerns legibility. Correct attribution values for right-to-left languages can be obtained from SHAP and LIME, but the resulting plots are not necessarily readable. The labels for the features may be misaligned, letters may be separated and the layout of the plot follows left-to-right conventions regardless of the reading direction. In these plots, explanation exists, but the reader cannot easily understand it. The challenge is not in the attribution values and/or the model itself. It is derived from the way the results are presented, therefore the solution needs to be implemented at the visualization level.

The second concerns interpretation. Even when a plot is drawn properly, the plot will primarily display the words with the highest or lowest importance scores. For instance, if a content moderator reviews an Urdu post, they can view the most useful words that helped the model to make the decision. But the scores do not tell why those words were significant in that specific post. They also fail to reveal the relationship between the words in the pattern, or the meaning of the pattern for the classifier's decision. Feature importance is useful, but it does not provide the full explanation. This is even more critical if the explanation must be comprehended within the reader's context and language.

These two problems motivate the following research questions.

**RQ1**. How can the visualization layer of SHAP and LIME be modified so that attributions are displayed consistently with the reading direction and script behavior of RTL languages, without altering the underlying attribution values, and how can the correctness of the resulting rendering be measured?

**RQ2**. How can attribution output be verbalized as a short contextual explanation in the reader's own language while remaining constrained to the features the attribution method identified?

## 5. Proposed solution

To address the problems identified, we have modified the original SHAP and LIME plotting functions and added a language-model-based explanation component. For RQ1, we have modified the SHAP plotting functions to support right-to-left languages. The modified version handles text direction and letter shaping, the first two of the three mechanisms identified in Section 3.1. Layout asymmetry is discussed in Section 8 and is left to future work. For RQ2, we added a language-model stage that verbalizes the attribution output in the reader's language; it is described in the final part of this section. Instead of showing only numerical attribution values, the system provides a simple explanation of the model's decision in the user's respective language. This helps users who may not be familiar with SHAP, or technical model outputs understand why certain words contribute to the prediction.

Our proposed package, named SHAP-RTL, addresses RQ1 by modifying the rendering stage of SHAP and LIME while keeping the model, tokenizer, explainer, and Shapley-value computation unchanged. Therefore, the attribution values, feature ordering, and clustering structure produced by standard SHAP remain unchanged. The package keeps the original plot-function interfaces, so existing code can be adapted by changing the import. RTL support is detected automatically. Labels containing RTL characters are processed through the RTL rendering pipeline, while labels without RTL characters follow the standard SHAP path. This also allows Urdu or Arabic text to appear alongside English text without applying the same direction to all labels.

The package uses two text-rendering paths. For general RTL text, it uses the Unicode Bidirectional Algorithm through python-bidi [20] and arabic-reshaper [21]. This approach is suitable for console output and scripts such as Hebrew and standard Arabic text. However, it is not sufficient for complex Urdu rendering, particularly Nastaliq, because its letter connections and positioning depend on font-specific OpenType rules. For plot labels, we therefore use HarfBuzz [22] and FreeType [23]. HarfBuzz performs bidirectional ordering, contextual shaping, and ligature substitution using the selected font, while FreeType rasterizes the resulting glyphs. The rendered text is placed as an image on the plot at the position of the original Matplotlib label. This approach preserves the original plot layout while replacing only the problematic text rendering. Because rendered labels are images rather than text, they cannot be searched or read by screen readers; Matplotlib 3.11.0's native path preserves these properties for Arabic and Persian. If HarfBuzz or FreeType is unavailable, the system falls back to PIL-based rendering. PIL does not join Arabic-script letters, so fallback output is comparable to the unmodified baseline; the package warns the user when the fallback is used.

An alternative would be to replace the plotting backend rather than the label renderer. The mplcairo backend routes text through Cairo and can perform complex text layout when compiled against Raqm, but that support is an optional build feature and is absent from the standard binary distributions, so a backend swap does not by itself enable shaping. A backend change also applies to the whole figure rather than to the labels that require it, and it addresses neither per-language font selection nor layout geometry. The rendering layer described here was therefore implemented at the label level, where the defect occurs and where the correction can be applied without altering the surrounding plot.

The choice of fonts is made based on the target language, and not just based on script. For Urdu the default is Noto Nastaliq Urdu, released under the SIL Open Font License; Nafees Nastaliq and Jameel Noori Nastaliq may be supplied by the user where licensing permits. Arabic is rendered with Noto Naskh Arabic, and Persian with Noto Naskh Arabic or Vazirmatn. Hebrew is rendered with Noto Sans Hebrew or Noto Serif Hebrew. Language-specific selection is required as Urdu, Arabic and Persian share the Arabic Unicode block but differ in typographic requirements.

The package provides RTL-aware versions of the SHAP plot functions listed in Table 1, namely summary, bar, beeswarm, heatmap, waterfall, force and decision, together with the LIME bar plot. Note that the summary plot is the legacy alias for beeswarm and dependence plot for scatter, so these are not counted separately. Both local and global explanations are supported. The bar plot also preserves SHAP's hierarchical clustering and dendrogram functionality.

For RQ2, we have added a language-model-based explanation stage to make the SHAP results easier for non-expert users to understand. The stage receives the source text, predicted label, ordered features, and their signed attribution values, and produces a short explanation in the target language. Thus, a user examining Urdu text can receive the explanation in Urdu rather than having to interpret numerical SHAP values alone. The language model is openai/gpt-oss-120b, accessed through the Groq API with a temperature of 0.1. The model is instructed to explain the provided attribution results rather than change, re-rank, or add features. Therefore, the SHAP values remain the basis of the explanation. The generated text should be treated as a user-friendly interpretation of the attribution results, not as independent evidence about the classifier.

## 6. Case study and experimental setup

To validate it we apply it to all four languages under study: Urdu, Arabic, Hebrew and Persian. For each we work with a dataset built around a socially relevant text classification task, hate speech or offensive language detection, since this is where post hoc explanation matters most in practice. A moderator reviewing a flagged post needs more than a label; they need to see which words led to that decision, and that explanation is only useful if it can be read in the order the post was written.

### 6.1 Datasets

The following benchmark datasets are used to evaluate the RTL-aware SHAP and LIME package. **Table 3** summarizes their composition. Counts are computed directly from the data files rather than quoted from the originating publications, and reflect the label configuration actually trained on.

1. **Urdu:** We have used NUHONS dataset [24], that consists of 16,337 cleaned Urdu YouTube comments, classified as 7,891 normal, 6,199 offensive, and 2,247 hate instances. The corpus authors report a Fleiss' Kappa of 0.653 across six annotators after calibration. Urdu is the primary case study because Nastaliq presents the most challenging shaping and rendering issues.
2. **Arabic: L-hsab** [25] dataset is used for testing on Arabic text. The dataset contains 5,847 Levantine Arabic tweets classified as normal, abusive, or hate. We merge abusive and hate into a non-normal class, resulting in 3,651 normal and 2,196 non-normal instances. The dataset reflects informal SM text, including code-switching and spelling variation.
3. **Hebrew:** For Hebrew, we have used the dataset accompanying the offensive language taxonomy proposed by Hamad et al. [18], The full corpus contains 15,881 tweets annotated by Arabic–Hebrew bilingual annotators into five categories: abusive, hate, violence, pornographic, and non-offensive. Following the original corpus authors, we combined the four offensive categories into a single offensive class. Our binary setup therefore contains 5,431 offensive tweets (34.20%) and 10,450 non-offensive tweets (65.80%), using the complete corpus rather than a subset. Hebrew serves as a control condition because its letters do not join, allowing directional and layout problems to be examined separately from shaping issues.
4. **Persian: PHate** [26] dataset is being used for testing on Persian text. This dataset contains 8,726 manually annotated Persian tweets. We use its aggregate HS label, comprising 4,866 positive and 3,860 negative instances. The dataset also provides human-annotated rationale spans, which are not used in the present study.

**Table 3:** Composition of the four benchmark corpora as used in this study.

| Language | Corpus | Source / platform | Instances | Task | Class distribution as used |
|---|---|---|---|---|---|
| Urdu | NUHONS [24] | YouTube comments (Nastaliq) | 16,337 | 3-class | Normal 7,891 (48.3%); Offensive 6,199 (37.9%); Hate 2,247 (13.8%) |
| Arabic | L-hsab [25] | Twitter (Levantine dialect) | 5,847 | Binary | Normal 3,651 (62.4%); Toxic 2,196 (37.6%) — abusive and hate merged |
| Hebrew | Hamad et al. [18] | Twitter | 15,881 | Binary | Non-offensive 10,450 (65.80%); Offensive 5,431 (34.2%) |
| Persian | PHate [26] | Twitter | 8,726 | Binary (from multi-label) | Hate speech 4,866 (55.8%); Normal 3,860 (44.2%) |
| PHate additionally carries independent labels for violence (783 positive), hate (2,475) and vulgarity (2,366), together with annotator rationale spans; we train against the aggregate hate-speech flag. | | | | | |

Together these four datasets span two scripts, two language families and three annotation schemes. Urdu and Persian share the Arabic script but differ in typographic tradition, Nastaliq against Naskh. Arabic contributes the same Naskh tradition in a dialectal register. Hebrew contributes a right-to-left script without cursive joining, which is what allows the directional and layout failure modes to be observed in isolation from the shaping failure mode. The corpora also differ in scale, from under six thousand instances to more than sixteen thousand, and in label structure, from binary to three-class to multi-label. This spread is what supports the claim that the alignment problem is a property of how the writing system is rendered rather than of any particular corpus, task or annotation scheme.

### 6.2 Classifier and experimental configuration

For testing our package on every dataset, we have used a TF-IDF vectorizer over word unigrams and bigrams, followed by LR with balanced class weights, applied one-vs-rest for the three-class Urdu corpus. Data is divided by an 80:20 stratified split preserving class proportions, with the random seed fixed at 42 throughout. SHAP and LIME explanations are then generated for up to 300 correctly and incorrectly classified instances, and the default visualization output is compared against the RTL output from our framework. The use of a simple linear classifier is deliberate because this study focuses on the explanation and rendering process rather than classification performance. A TF-IDF-based linear model provides direct feature-to-token correspondence, allowing each SHAP value to be linked to an identifiable term in the source text. It also enables efficient and reproducible attribution computation, ensuring that differences between the original and modified plots are due to rendering rather than changes in the underlying values.

### 6.3 Classifier performance

The classifiers provide a substrate for generating attributions and are not a contribution of this work. The rendering results in Section 7 are independent of classifier quality, since a label is rendered correctly or incorrectly irrespective of whether the underlying prediction is right. Performance is reported in Table 4 so that readers can judge what the attributions describe.

**Table 4:** Performance of the TF-IDF and logistic regression classifiers used to generate the attributions. The majority baseline is the accuracy obtainable by always predicting the most frequent class.

| Language | Classes | Train acc. | Test acc. | Macro F1 | Majority baseline |
|---|---|---|---|---|---|
| Urdu | 3 | 0.697 | 0.629 | 0.580 | 0.483 |
| Arabic | 2 | 0.907 | 0.801 | 0.786 | 0.624 |
| Persian | 2 | 0.799 | 0.738 | 0.736 | 0.558 |
| Hebrew | 2 | 0.727 | 0.684 | 0.702 | 0.658 |

**Table 4** shows the performance of the classifiers used to generate the SHAP values. The Urdu classifier reached 0.629 test accuracy, compared with a majority-class baseline of 0.483. Its Macro F1 was 0.580. The Arabic classifier performed better, with 0.801 test accuracy and a Macro F1 of 0.786, above its majority baseline of 0.625. The Persian classifier also performed above its majority baseline, with 0.738 test accuracy compared with 0.558, and a Macro F1 of 0.736. The Hebrew classifier reached 0.684 test accuracy against a majority-class baseline of 0.658, with a Macro F1 of 0.702. Because all four classifiers are trained with balanced class weights, which raise recall on the minority classes at the cost of majority-class accuracy, Macro F1 gives the clearer comparison against a trivial predictor: a majority-class predictor scores 0.397 Macro F1 on the Hebrew test split and 0.217 on the three-class Urdu split. Every classifier exceeds its trivial counterpart on both measures.

### 6.4 Evaluation protocol for rendering correctness

Rendering correctness is assessed by an OCR round trip. Each feature label is rendered as it would appear on an attribution plot, read back with Tesseract 5.5.3 [27] using the language model matching the label's script in single-word page segmentation mode, and compared against the source string. The metric is CER, the Levenshtein distance between source token and recovered string normalized by source length. The premise is that a label is correctly rendered if the token can be recovered from the rendered pixels: CER near zero indicates that the glyphs carry the token, and CER near one indicates that they do not. Before comparison, both strings are NFC-normalized and stripped of bidirectional control characters, zero-width joiners and tatweel, none of which are visible on the page and none of which therefore constitute rendering failures.

The evaluation covers 200 feature words per language. Words are drawn from the pooled top-five attributions of up to 300 explained test instances, which is the set that appears on the figures, and are stratified across three string-length bands so that the sample is not dominated by short high-frequency terms. Multi-word bigram features are retained, since joining behavior across an intervening space is itself a rendering concern.

Four conditions are compared. Condition A passes the label unmodified to the plotting layer with a script-appropriate font, reproducing default behavior. Condition B applies contextual reshaping followed by reordering under the UBA [20], the prevailing community workaround. Condition C is a reference CTL pipeline using HarfBuzz shaping [22] and FreeType rasterization [23], implemented independently of the proposed package. Condition D is SHAP-RTL. A same-pipeline render at larger scale provides a OCR reference on the recognition accuracy achievable for each language, which is necessary because Tesseract is not equally accurate across scripts and a residual error rate cannot otherwise be attributed.

All conditions are measured on identical hardware, fonts and OCR configuration under Matplotlib 3.10.8 and 3.11.0, with library version as the only manipulated variable. Confidence comparisons use a paired bootstrap on the same word set, since per-word CER is bounded and right-skewed and a parametric interval would be inappropriate. Fonts are Noto Naskh Arabic for Arabic and Persian, Noto Nastaliq Urdu for Urdu and Noto Sans Hebrew for Hebrew, all under the SIL Open Font License, held constant across conditions, so font selection itself is not measured. Exact package versions are listed in the repository's requirements file.

## 7. Results

This section presents the results through paired comparisons of the default SHAP or LIME plots and the corresponding plots produced by the proposed rendering layer. Each pair uses the same model, attribution values, and fixed random seed, so any differences between the plots are due to the rendering process. The results are discussed separately for each plot family, with the visual comparisons providing the main evidence for the proposed approach.

### 7.1 Rendering correctness

**Table 5** reports CER under Matplotlib 3.10.8 for 200 feature words in each language. Under default handling the defect is present in all four languages, with CER between 0.820 and 0.979. At these values the rendered label does not carry its token: a reader cannot recover the feature name from the figure, which is the condition described in Section 3 and illustrated qualitatively in Fig. 2 and the default panels of **Figs. 3–10**.

**Table 5**: Character error rate by condition under Matplotlib 3.10.8 (n = 200 words per language). Lower is better; 0.000 indicates perfect recovery.

| Language | A: unmodified | B: reshape + reorder | C: reference CTL | D: SHAP-RTL | OCR reference |
|---|---|---|---|---|---|
| Arabic | 0.820 | 0.011 | 0.012 | 0.013 | 0.021 |
| Persian | 0.838 | 0.010 | 0.006 | 0.009 | 0.019 |
| Hebrew | 0.842 | 0.131 | 0.135 | 0.136 | 0.236 |
| Urdu | 0.979 | 0.998 | 0.351 | 0.361 | 0.374 |

The community workaround resolves Arabic, Persian and Hebrew, bringing all three to within 0.01 of the OCR reference. For Urdu it fails, scoring 0.998, marginally worse than applying no correction at all; the paired bootstrap difference against full shaping is 0.646 with $p < 0.001$. The mechanism is identifiable from the rendering log. Contextual reshaping substitutes characters into the Unicode presentation forms blocks, and fonts such as Noto Nastaliq Urdu implement shaping through GSUB and GPOS tables rather than carrying those code points. Noto Nastaliq Urdu contains none of them, and the renderer reported missing-glyph errors for the majority of substituted characters. This confirms, with a mechanism, that contextual form substitution is insufficient for Nastaliq, where glyph placement is cascading and vertically stacked rather than linear.

SHAP-RTL reaches the OCR reference in every language and matches the independent reference implementation to within 0.010 CER in every cell. The agreement between conditions C and D is the expected outcome, since both perform complex text layout; it is reported as a correctness check on the implementation rather than as evidence of superiority over complex text layout as a technique. The contribution of the package lies in performing that layout automatically, with per-language font selection, inside the existing explanation workflow.

The residual error for Urdu at 0.361 and Hebrew at 0.136 is a property of the recognition engine and not of the rendering. The OCR reference for those languages is 0.374 and 0.236 respectively, indicating that recognition accuracy on Nastaliq and on short Hebrew tokens is limited independently of how the text is drawn. The metric therefore has less headroom for Urdu than for Arabic or Persian, and the Urdu result should be read as recovery to the recognition ceiling rather than to zero. That the OCR reference is itself slightly worse than conditions C and D for these two languages reflects the larger render scale used for the bound, to which the recognition engine is sensitive.

### 7.1.1 Behavior across library versions

**Table 6** repeats the evaluation under Matplotlib 3.11.0 on the same words, fonts, hardware and OCR configuration. The rewritten text pipeline in 3.11.0 resolves default rendering for Arabic and Persian and improves Hebrew and Urdu to near the OCR reference, confirming that the upstream change addresses directional ordering and contextual joining. The community workaround, by contrast, degrades under 3.11.0 across all four languages, to between 0.815 and 0.981. Because the workaround pre-reorders text into visual order, which the new bidirectional pass then reorders a second time, upgrading the plotting library silently invalidates code that adopted the workaround: output that rendered correctly before the upgrade becomes unreadable after it, with no error raised.

**Table 6**: Character error rate under Matplotlib 3.11.0 (n = 200 words per language).

| Language | A: unmodified | B: reshape + reorder | D: SHAP-RTL |
|---|---|---|---|
| Arabic | 0.011 | 0.839 | 0.013 |
| Persian | 0.010 | 0.815 | 0.009 |
| Hebrew | 0.147 | 0.838 | 0.136 |
| Urdu | 0.344 | 0.981 | 0.361 |

SHAP-RTL returns identical scores under both versions in all four languages, because label rasterization is performed independently of the plotting library's text engine. Of the approaches evaluated it is the only one correct under both versions, and this version independence is the practical argument for a dedicated rendering layer rather than reliance on either the default path or the community workaround.

### 7.2 Addressing right-to-left content in the SHAP method

The beeswarm results provide the first comparison in this set, as presented in **Fig. 3.** In the default Urdu and Arabic plots (a) and (c), the feature labels appear as disconnected glyphs and are displayed in the wrong direction.

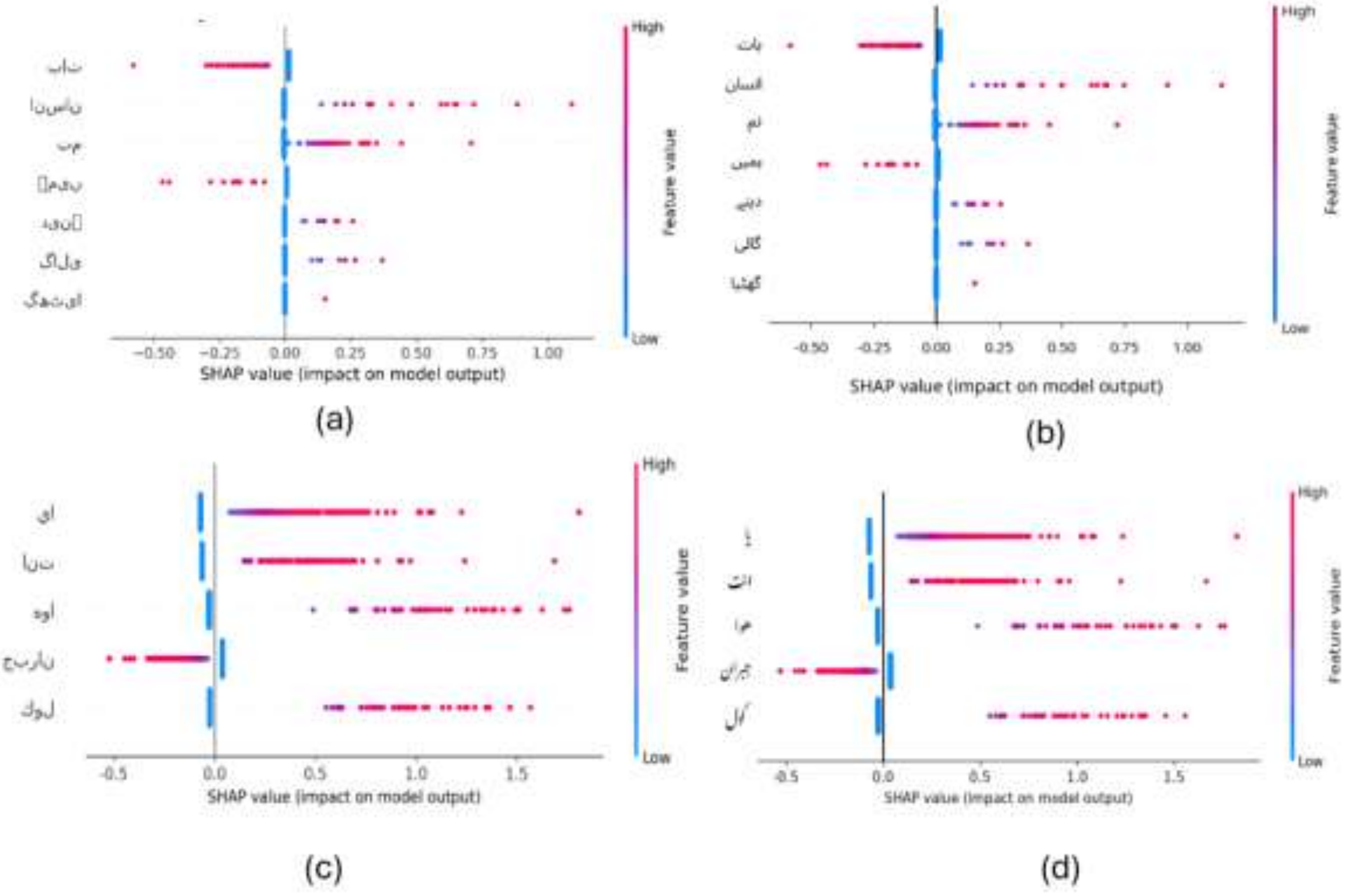

**Fig. 3:** SHAP Beeswarm comparison of standard versus RTL rendering: (a) misaligned Urdu, (b) RTL Urdu, (c) misaligned Arabic, (d) RTL Arabic.

Although individual letters may still be recognizable, the complete words are difficult to identify because connected letter forms are important for reading Arabic-based scripts. After applying the proposed rendering layer, panels (b) and (d) display the same features with correct shaping and ordering. The beeswarm plot also provides a useful test because the labels are positioned beside a dense set of individual points. Any mismatch between a label and its corresponding row would therefore be easy to notice, but no such problem is visible in the corrected plots.

A different pattern can be seen when Urdu and Hebrew are compared in **Fig. 4**. The two languages help separate the RTL problems identified earlier because Hebrew letters do not connect. In the default Hebrew plot (c), the labels are readable but appear in the wrong direction, indicating a directional problem without shaping failure. Urdu in panel (a), in contrast, shows both incorrect direction and incorrect letter shaping. The corrected plots reflect this difference: Hebrew mainly requires reordering, whereas Urdu also requires proper reconstruction of connected forms. This comparison supports the distinction between the different RTL rendering problems rather than treating them as one general issue.

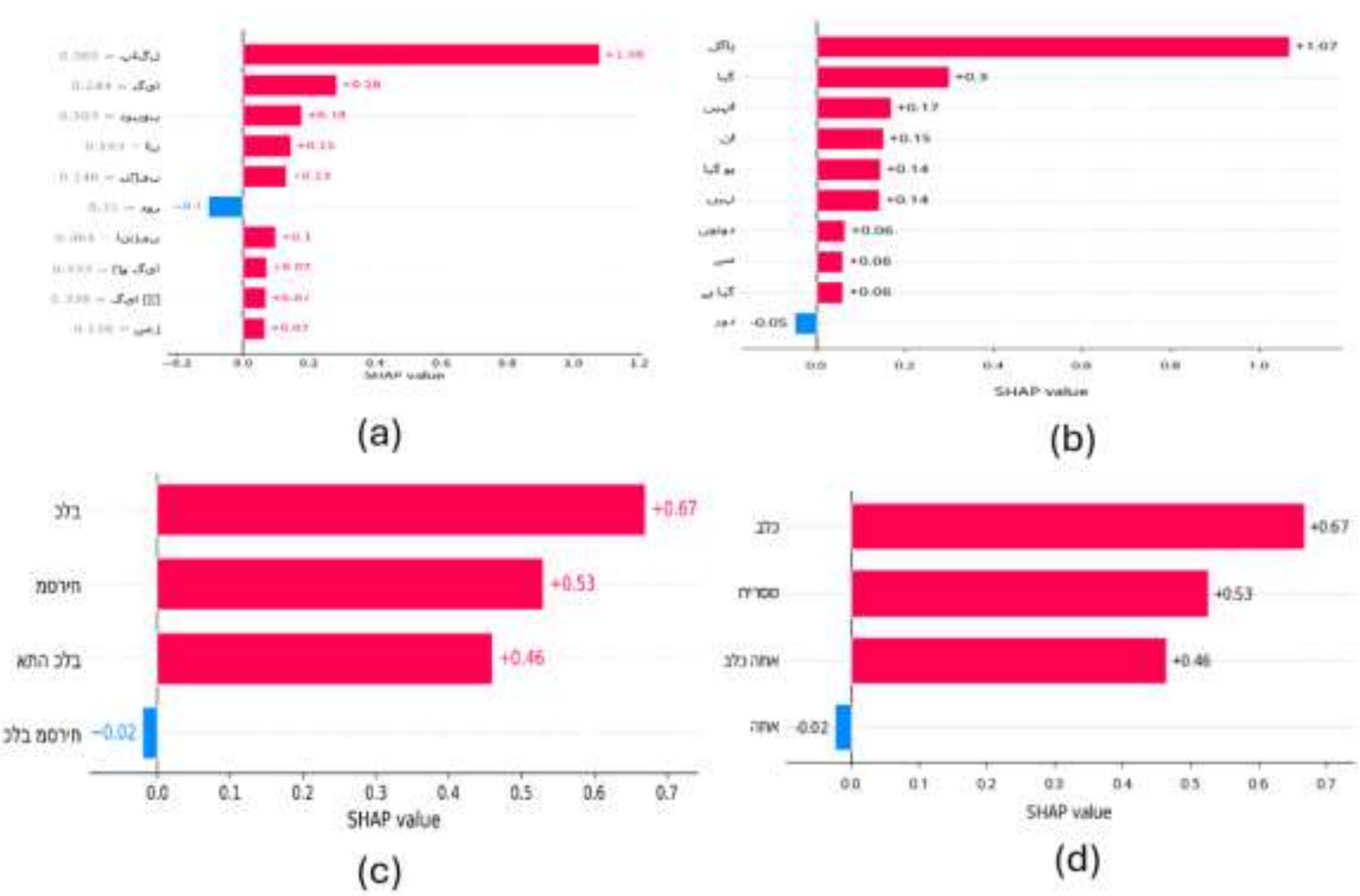

**Fig. 4**: SHAP Bar plot comparison of standard versus RTL rendering: (a) misaligned Urdu, (b) RTL Urdu, (c) misaligned Hebrew, (d) RTL Hebrew.

The waterfall results in **Fig. 5** provide a further test because this plot represents a local explanation for a single prediction. Each feature label is positioned next to its signed contribution and the corresponding cumulative value. The reader therefore needs to connect the feature, its contribution, and its direction at the same time. Incorrectly rendered labels make this association more difficult, while the corrected panels (b) and (d) make these elements easier to follow.

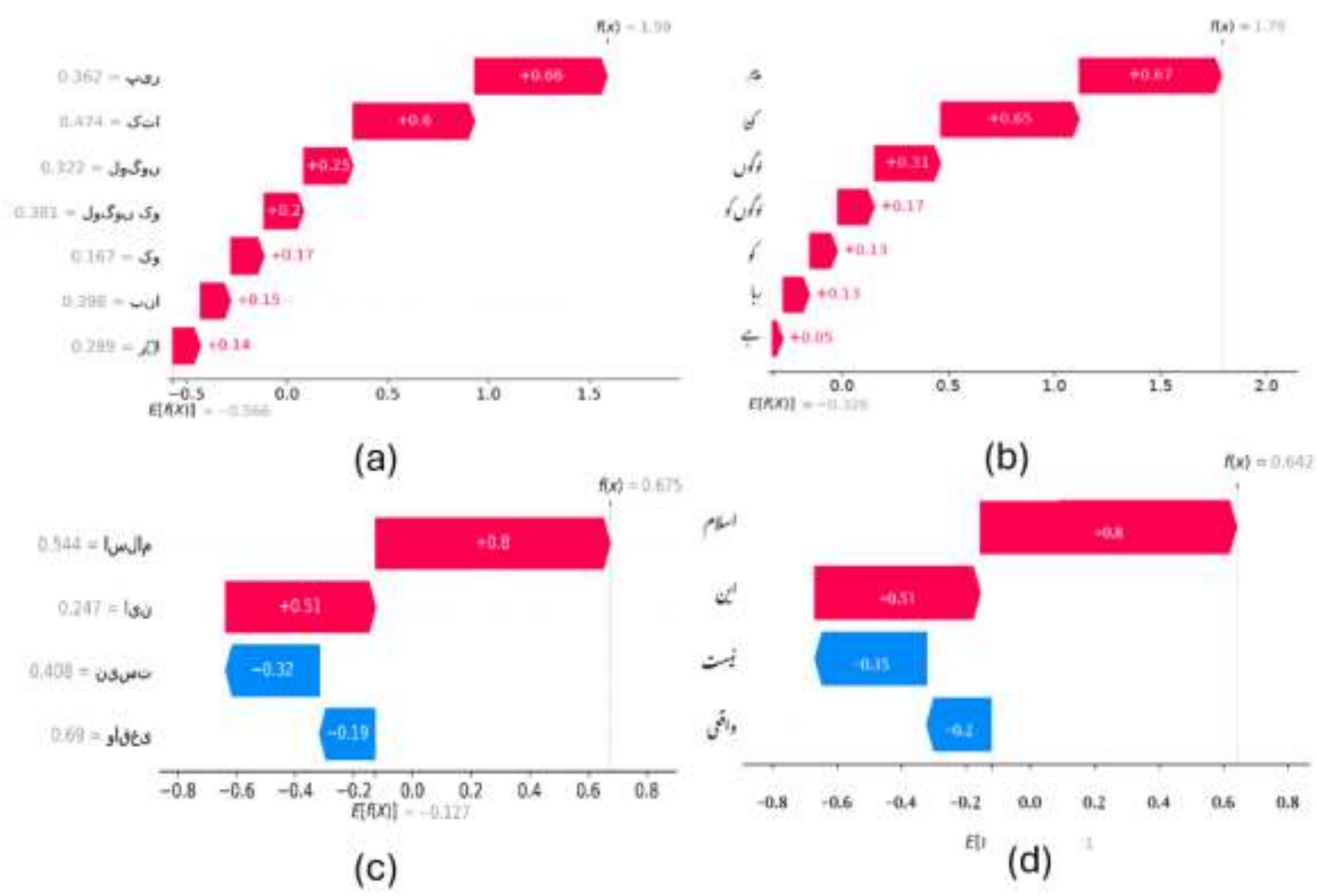

**Fig. 5**: SHAP Waterfall comparison of standard versus RTL rendering: (a) misaligned Urdu, (b) RTL Urdu, (c) misaligned Persian, (d) RTL Persian.

For the force plot, shown in **Fig. 6**, the rendering problem becomes more challenging. Unlike the previous plots, the force plot has a strong horizontal flow from the base value toward the final prediction. The proposed layer correctly renders the individual feature labels, making their contributions readable. However, the overall plot still follows a left-to-right visual flow, so the geometry itself has not been fully adapted for RTL reading.

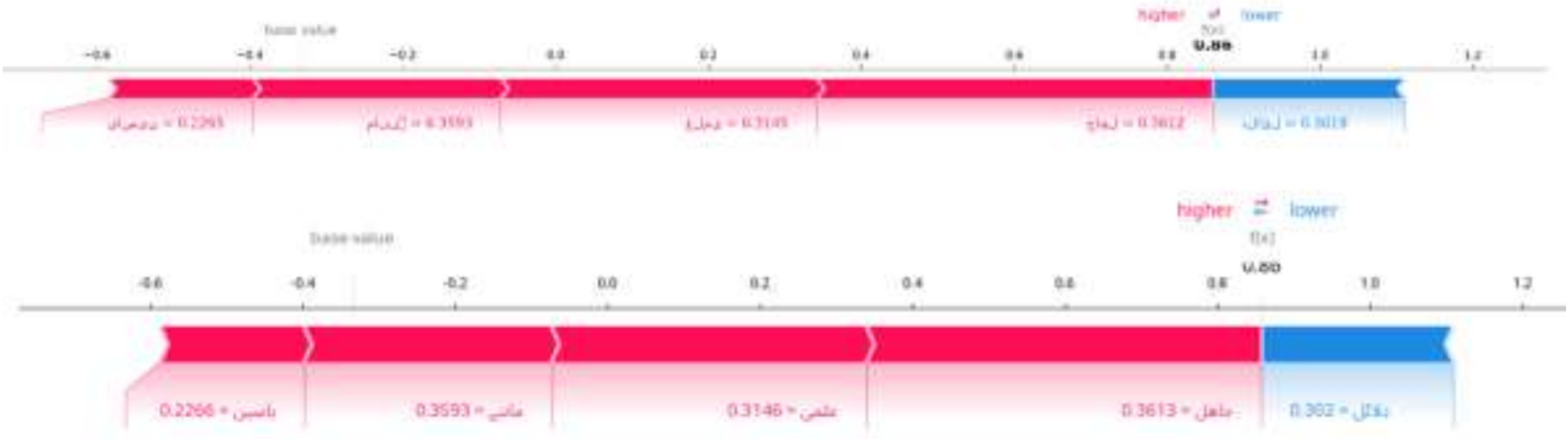

**Fig. 6**: SHAP Force plot comparison: (top) misaligned Urdu and (bottom) RTL Urdu rendering.

**Fig. 7** shows the decision plot for Arabic. The plot traces the cumulative contribution of ordered features along a common axis, leaving limited space between labels. When text is incorrectly rendered, it becomes difficult to identify the individual features from their position alone. The corrected panel shows that the proposed rendering approach can handle this dense arrangement without visible label collision or overlapping.

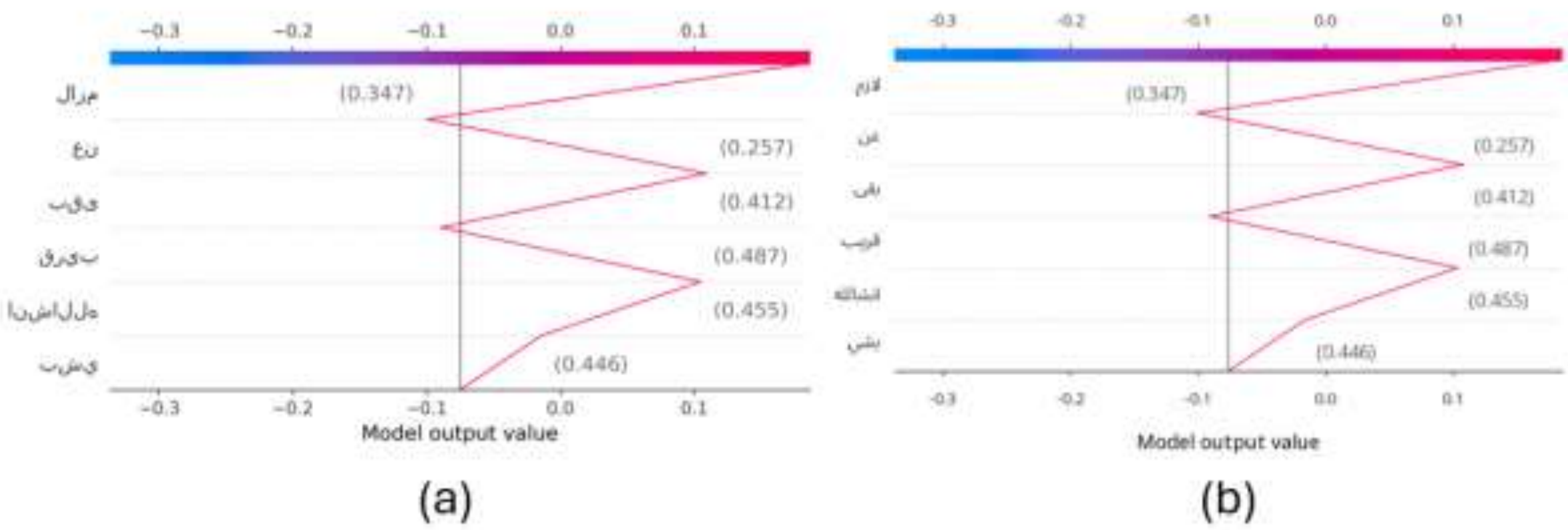

**Fig. 7**: SHAP Decision plot comparison showing (a) misaligned keywords and (b) RTL keywords in Arabic.

The heatmap in **Fig. 8** extends the evaluation to a global explanation. Here, feature labels are used as row identifiers across multiple instances, so a single unreadable label can affect the interpretation of a much larger pattern. The corrected heatmap keeps the feature names readable while preserving their position within the plot. Together with the beeswarm and Urdu-Hebrew comparisons, these results indicate that the proposed rendering approach works across both local and global SHAP explanations.

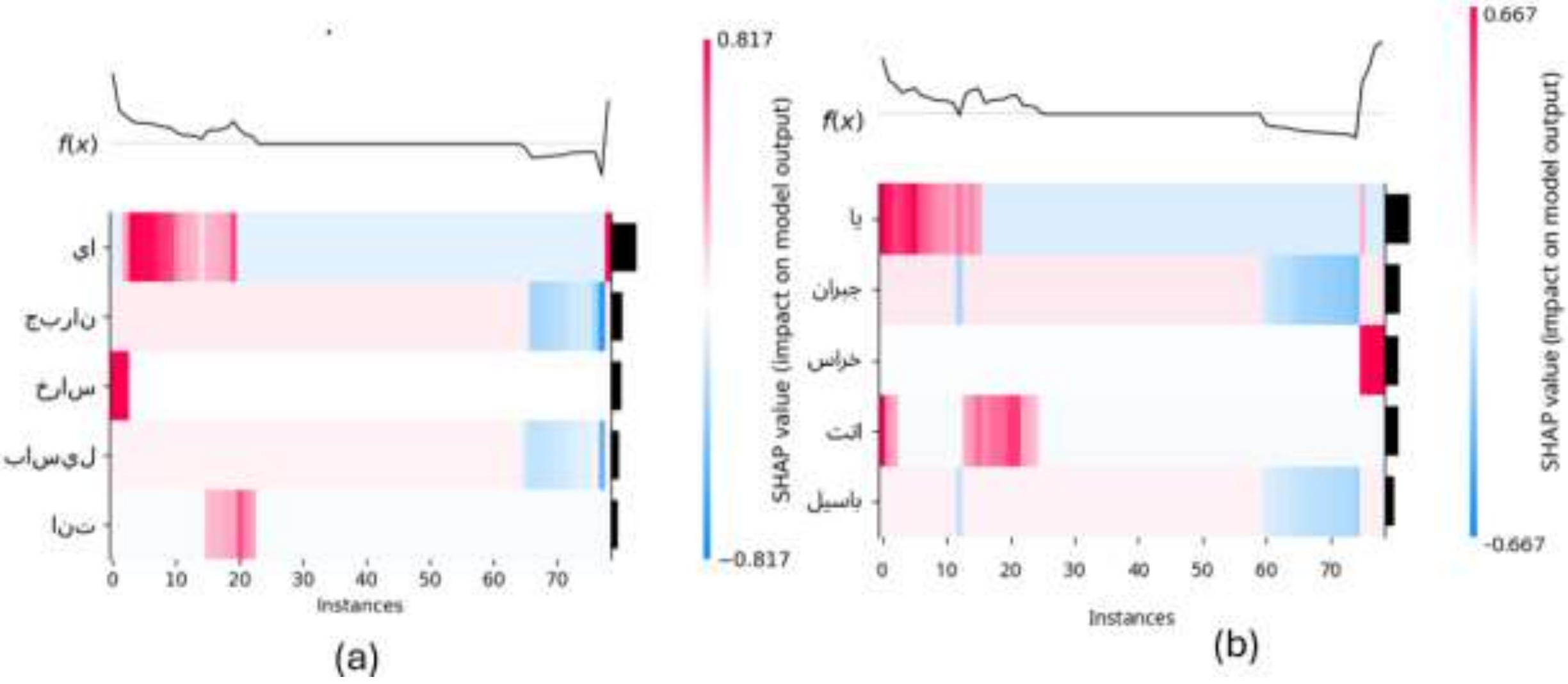

**Fig. 8**: SHAP Heatmap plot comparison showing (a) misaligned keywords and (b) RTL keywords in Arabic.

### 7.3 Addressing right-to-left content in the LIME method

The section examines LIME explanations for two languages: Urdu, and Arabic. This comparison is included to determine whether the RTL rendering problems observed with SHAP are specific to the explanation method or arise more generally from the visualization process. Since LIME uses a different approach to calculate feature importance, similar rendering behavior across both methods would provide stronger evidence that the problem occurs at the visualization level.

The LIME results for Urdu are presented in **Fig. 9**. The default plot shows the same type of text-order and shaping problems observed in the corresponding SHAP plot, while the proposed rendering layer produces correctly ordered and shaped labels. The similar results across the two methods indicate that the problem is not specific to SHAP.

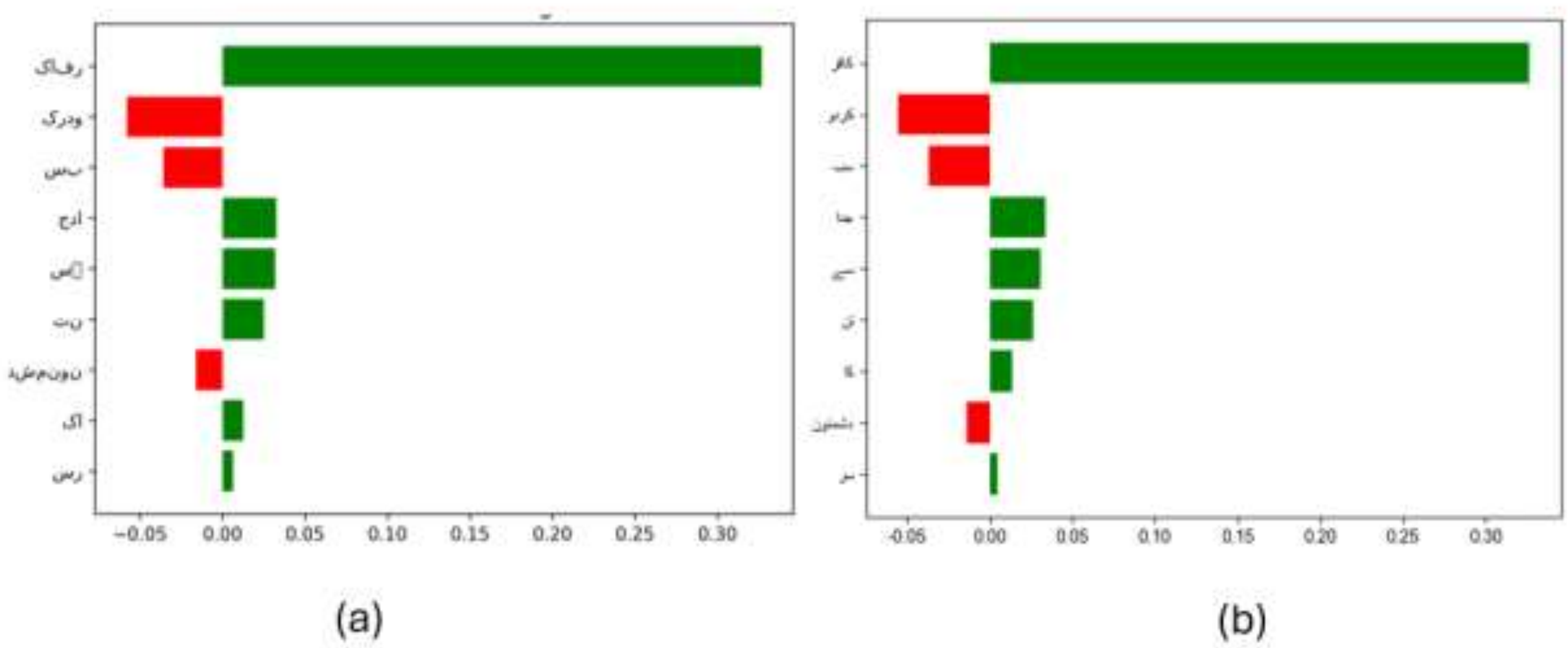

**Fig. 9**: LIME Bar plot showing (a) misaligned keywords and (b) RTL keywords in Urdu.

A similar pattern is observed for Arabic in **Fig. 10**. LIME uses a different explanation approach based on local linear models rather than Shapley values, yet the same RTL rendering problems appear in its default output. This provides further evidence that the issue lies mainly in the visualization layer rather than in the underlying explanation method. It also suggests that the proposed rendering approach could be adapted to other explainability tools that rely on the same visualization framework.

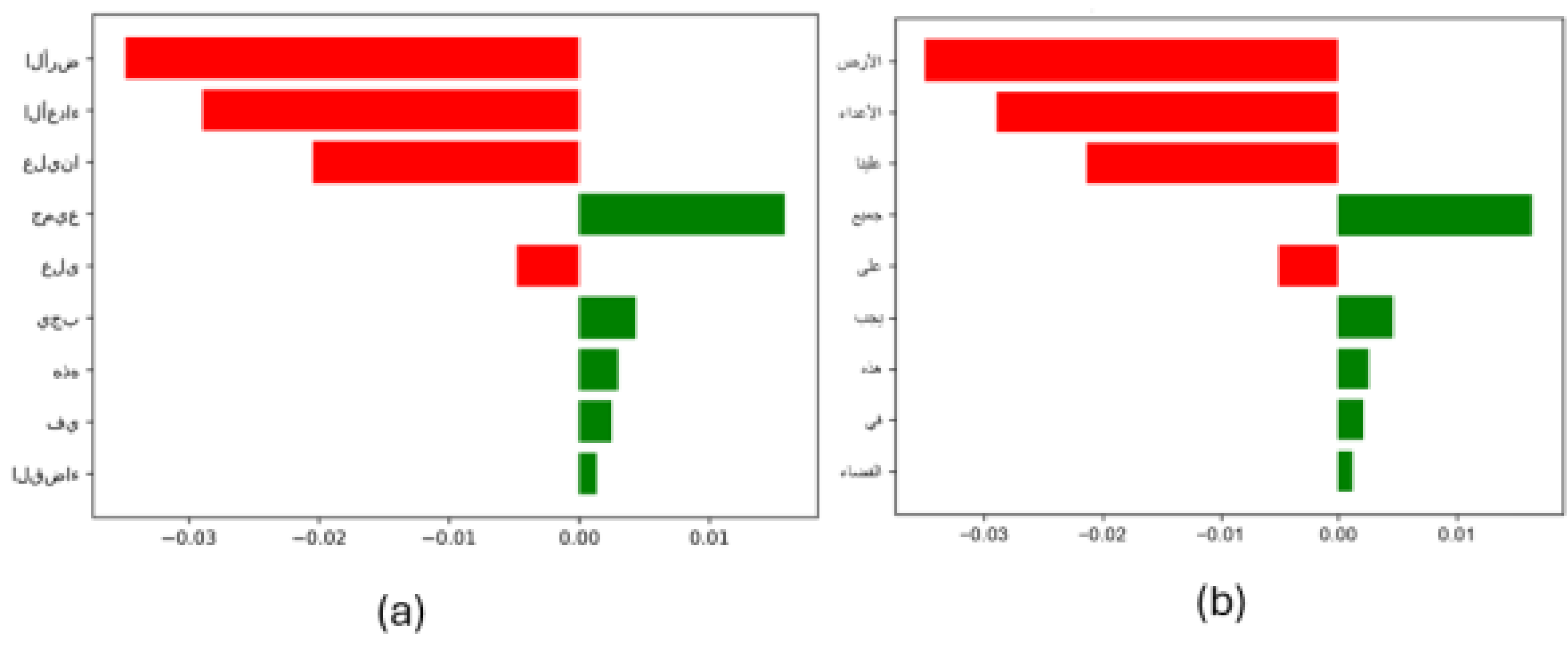

**Fig. 10**: LIME bar plot showing (a) misaligned keywords and (b) RTL keywords in Arabic.

### 7.4 Contextual explanation stage

The explanation component follows a five-stage workflow, allowing users to control the level of detail while keeping the same SHAP and LIME values throughout.

1. **Rendering:** The SHAP/LIME plot is generated in the target language using the RTL pipeline.

2. **Feature selection:** Users choose how many top features to explain as shown in **Fig.11**. Zero or an empty entry skips this step, while values above the available features are automatically limited.

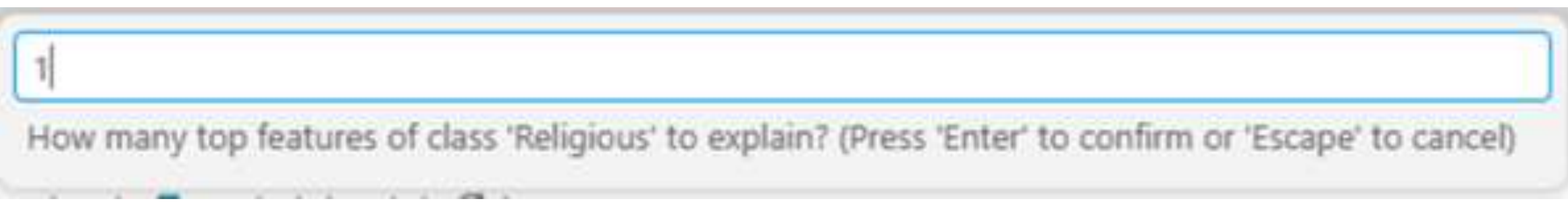

**Fig. 11:** Feature-count prompt after plot output (1st prompt)

3. **Target-language explanation:** The selected features, attribution values, source text, and predicted label are passed to the language model to generate a short explanation in the target language as shown in **Fig. 12**.

Analysis of Features for Religious (Urdu)

لفظ: کافر: [لفظ "کافر" کا استعمال مذہبی سیاق و سباق میں ہوتا ہے، خاص طور پر جب کسی کو غیر مسلم یا مذہب کے مخالف کے طور
پر پیش کیا جاتا ہے۔ اس جملے میں "سب کافر دشمنوں کا سر تن سے جدا کر دو" میں لفظ "کافر" ان لوگوں کی طرف اشارہ کرتا ہے جو
مقررہ مذہبی عقائد کے خلاف ہیں یا ان کے پیروکار نہیں ہیں۔ یہ لفظ مذہبی بنیاد پرستی اور شدت پسندی کی عکاسی کرتا ہے، جہاں
مخالفین کو مذہبی دشمن کے طور پر دیکھا جاتا ہے اور ان کے خلاف تشدد کی ترغیب دی جاتی ہے۔ اس طرح، یہ لفظ مذہبی تناظر میں
شدت پسندی اور عدم برداشت کی نشاندہی کرتا ہے۔]

**Fig. 12:** Explanation of feature in respective language

4. **English plot:** Users can optionally generate the same plot with English feature labels by saying yes to the prompt as shown in **Fig. 13**. The attribution values remain unchanged.

Do you want English Plot and Analysis? (yes/no): (Press 'Enter' to confirm or 'Escape' to cancel)

**Fig. 13**: Asking the user for English plot and analysis

5. **English explanation:** Users can also request an English explanation, with the number of features controlled separately as shown in **Fig. 14**.

```
English Translation: The infidel's death has come, kill this traitor now.
--------------------------------------------------------------------------------
Top Features for Religious (English Analysis)
Word: Infidel: The word "کافر" translates to "infidel" in English and is often used in a religious context to denote someone
who does not believe in a particular faith, especially Islam. In the given translation context, the term is used pejoratively
and is associated with violence and hostility, which can trigger the model's 'Religious' classification. The use of "کافر" in
this context suggests a religiously motivated conflict or animosity, as it labels someone as an outsider or enemy based on
their religious beliefs. This classification is further reinforced by the surrounding language that incites violence,
indicating a potential threat or extremist sentiment rooted in religious intolerance.
```

**Fig. 14**: English-language explanation (NLP based)

The target-language output supports native readers, while the English output provides an accessible version for reviewers, collaborators, and other users. Both are generated from the same attribution values.

**Figs. 15-18** show the output of the contextual explanation stage for Urdu, Hebrew, Persian and Arabic. In each case the stage receives the corrected attribution set together with the source

text and returns a short account in the target language, identifying which terms drive the classification and how they function in context.

The Urdu example in **Fig. 15** illustrates the explanation stage. While the SHAP values identify an important term, the generated explanation provides its context by showing how the term becomes offensive through its surrounding language. This contextual information is not captured by the attribution values alone.

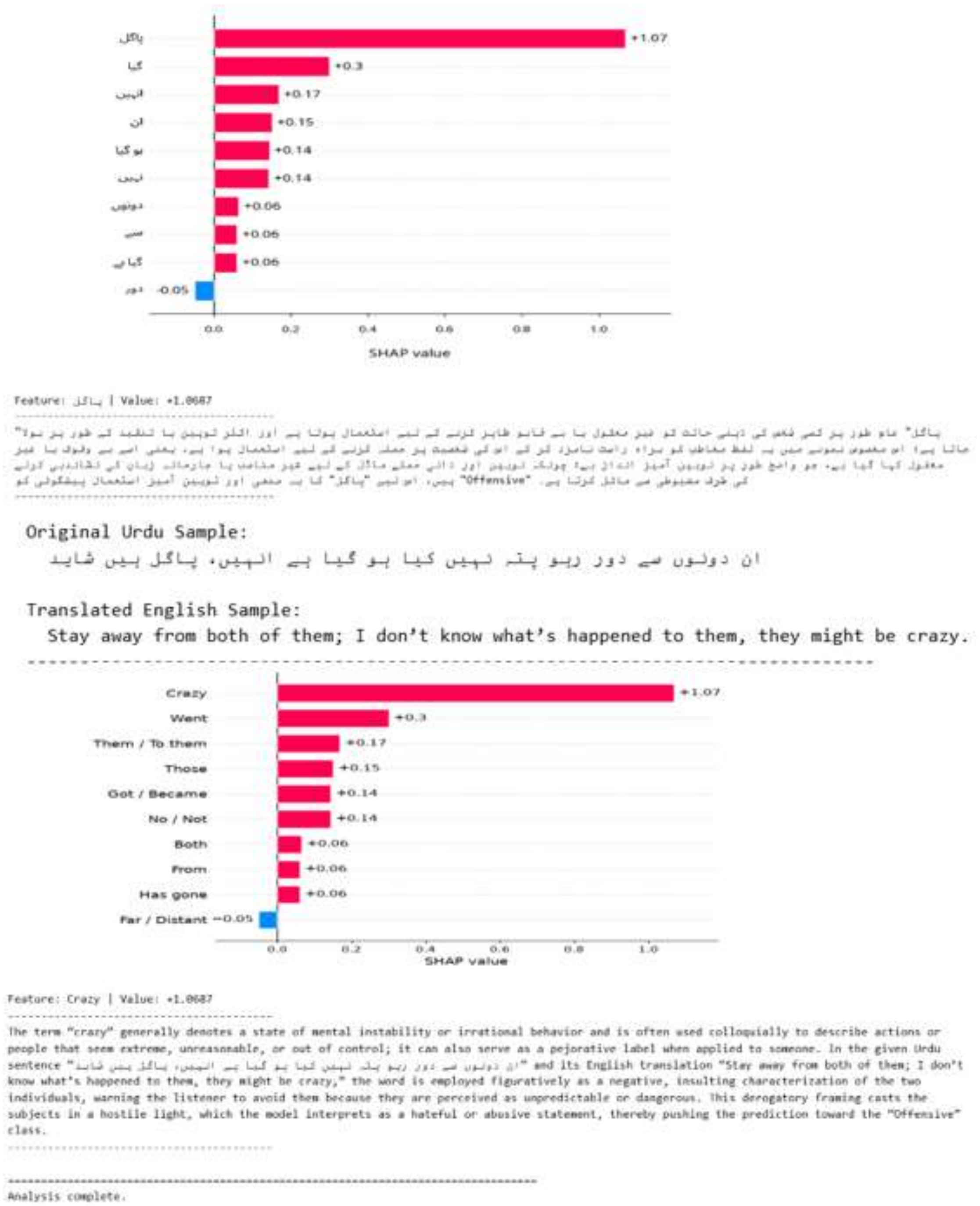

**Fig. 15**: Language-model contextual explanation for Urdu.

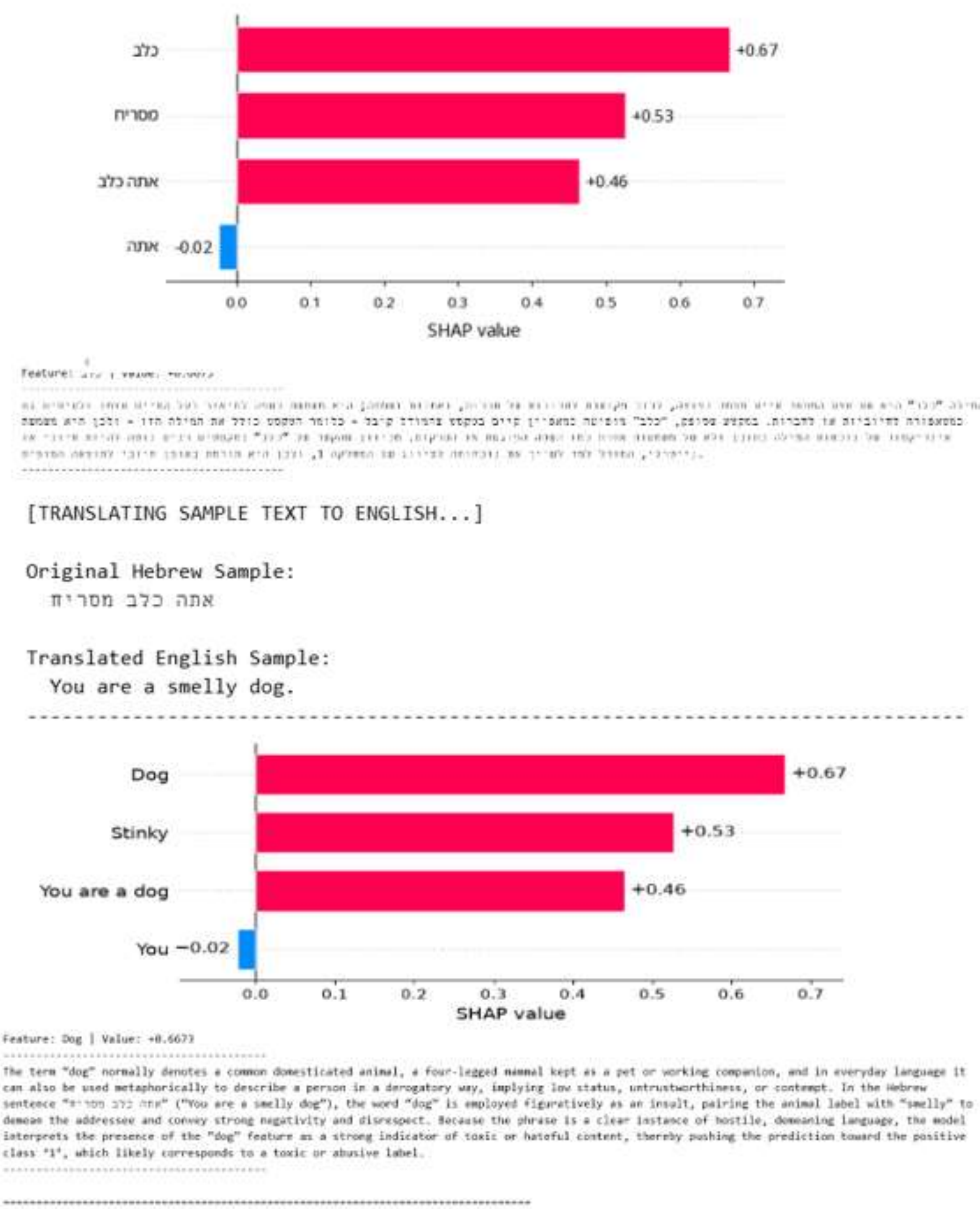

**Fig. 16**: Language-model contextual explanation for Hebrew

## 8. Conclusion

Attribution plots for right-to-left text render labels that readers cannot recover, while the attribution values behind them stay correct. The failure lies in how the labels are drawn, and it separates into direction, shaping and layout. This paper built a rendering layer that intercepts label drawing in SHAP and LIME and measured it by an OCR round trip over 200 feature words per language. Under Matplotlib 3.10.8 the default output reached CER of 0.820 for Arabic, 0.838 for Persian, 0.842 for Hebrew and 0.979 for Urdu. The rendering layer brought all four to the OCR reference level.

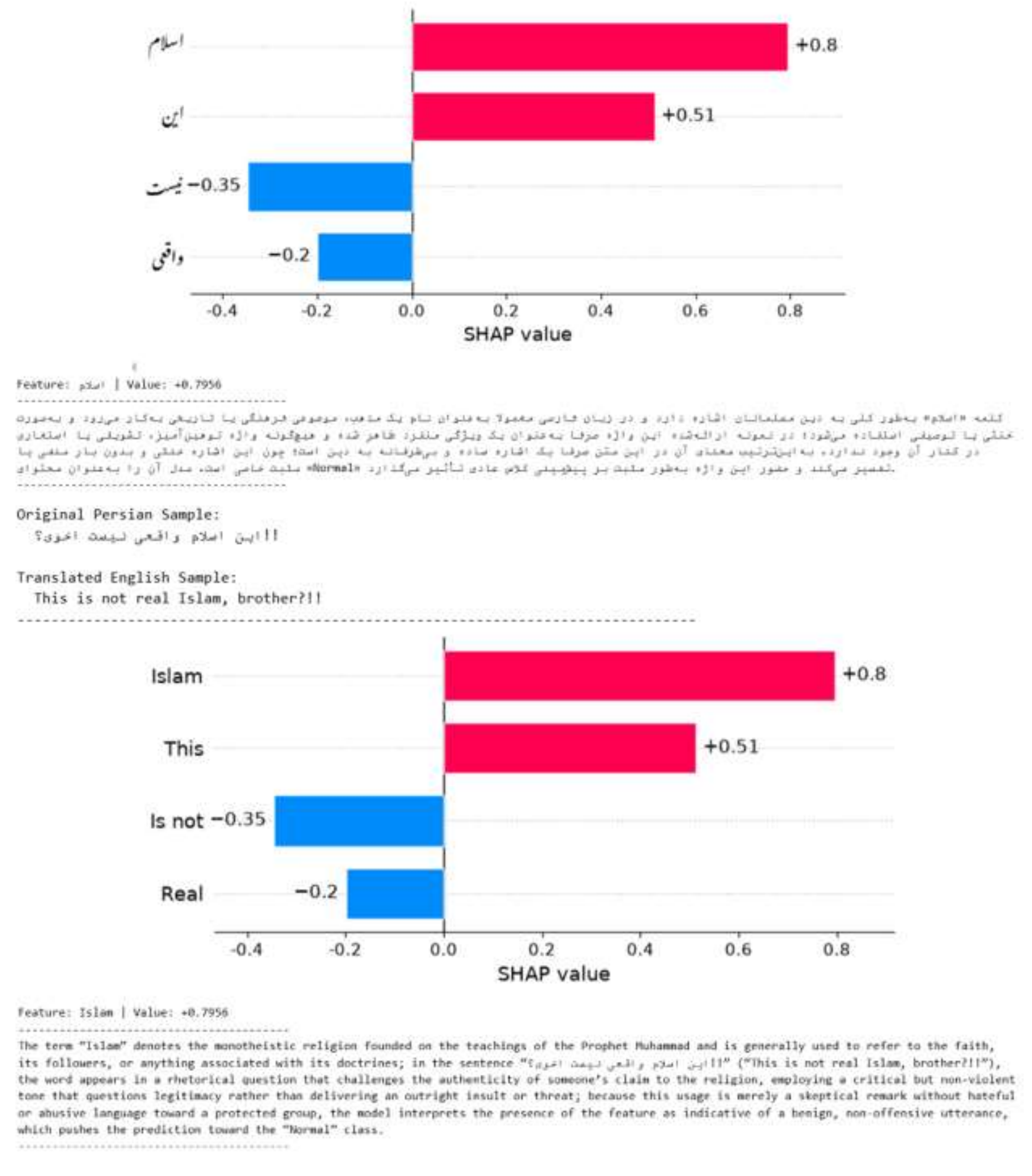

**Fig. 17**: Language-model contextual explanation for Persian

Two findings extend past the package. Contextual reshaping followed by bidirectional reordering, the workaround usually recommended for this problem, fails for Urdu at 0.998, worse than leaving the label untouched, because the presentation forms it substitutes are absent from Noto Nastaliq Urdu. That workaround also degrades across all four languages under Matplotlib 3.11.0, whose new bidirectional pass reorders the pre-reordered text a second time, so code adopting it will break on upgrade with no error raised. The rendering layer scores identically under both versions, since it rasterizes labels outside the plotting library's text engine.

Layout asymmetry remains unresolved and plot geometry still flows from left to right. The OCR metric measures machine recoverability, so a study with native speakers is needed to establish improved comprehension. Word-level tokenization is a precondition, since a sub-word fragment of an Arabic-script word has no valid isolated form. The contextual explanation stage has not yet been evaluated for faithfulness.

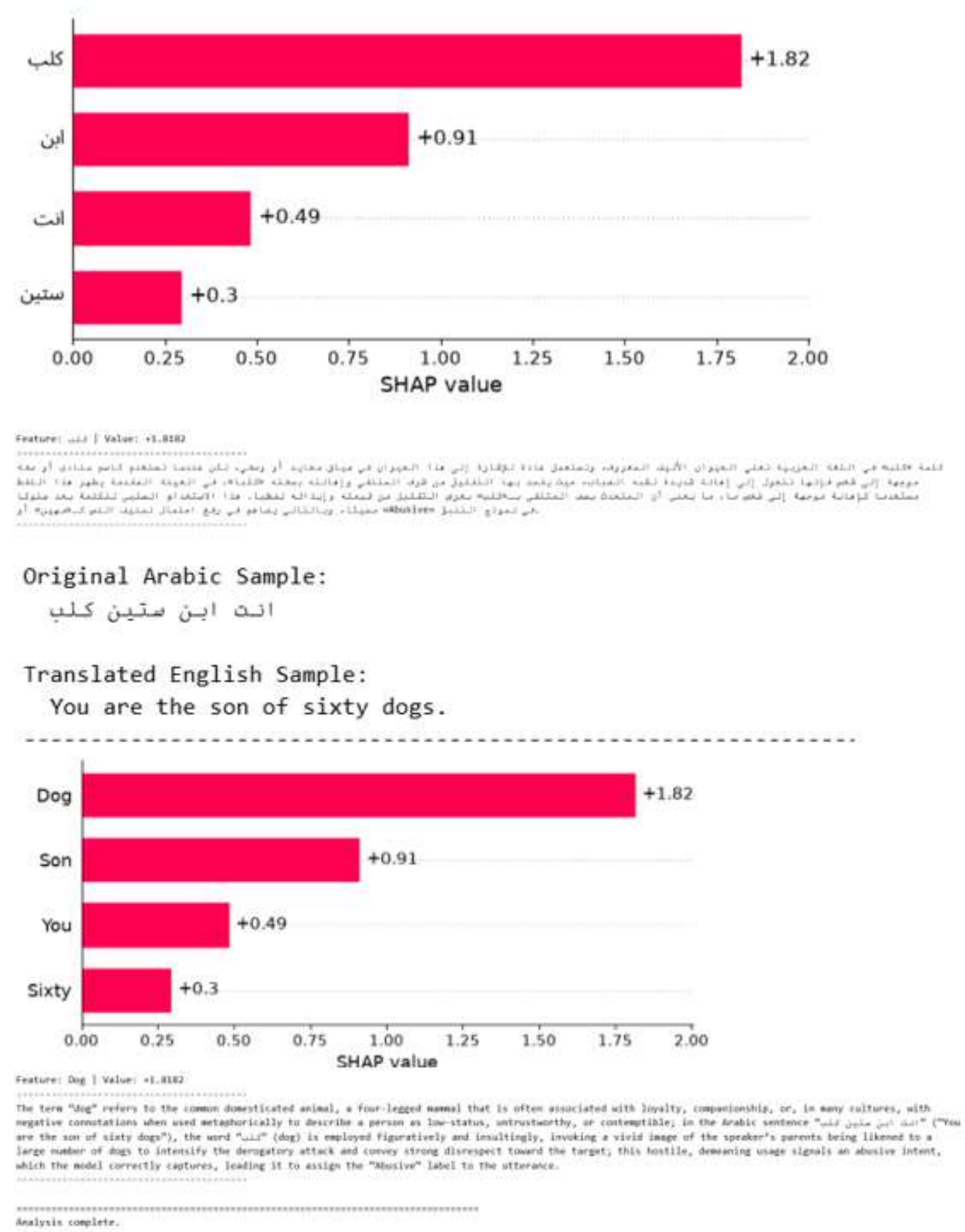

**Fig. 18**: Language-model contextual explanation for Arabic.

## Ethics Statement

The four corpora used in this study consist of publicly posted SM text annotated for hate speech or offensive language by their original authors, and are used under the terms those authors specify. The annotation labor is theirs and is credited in Section 6.1. Several figures reproduce offensive terms drawn from these corpora, which is necessary to demonstrate that the rendered labels are legible; readers should be aware of this before viewing **Figs. 2 to 18**.

The contextual explanation stage transmits the source text and the attribution set to a third-party language model API. In a deployed moderation setting this would mean sending user-generated content to an external service, and an operator handling personal data would need either a locally hosted model or an agreement covering that transfer. The stage is optional and the rendering layer functions without it.

The explanations generated by that stage are fluent by construction and may read as authoritative whether or not the attributions they describe are faithful to the model. The stage is instructed to describe the supplied attribution set rather than to reason independently, and its output should be treated as a presentation of those attributions rather than as evidence about the classifier.

**Author Contributions**

Rameesha Zia: Visualization, Validation, Software, Investigation, Formal analysis, Data curation, Writing – original draft, Muhammad Shahid Iqbal Malik: Conceptualization, Methodology, Supervision, Formal analysis, Validation, Data curation, Writing – original draft, Writing – review & editing

**Funding**

This research did not receive any specific grant from funding agencies in the public, commercial, or not-for-profit sectors.

**Data and code availability**

The SHAP-LIME-RTL package, including the rendering layer, the plot implementations and the explanation stage, are available at:

https://github.com/MSIMALIK/Ext-SHAP-LIME-Urdu, and
https://github.com/Ramishazia01/Ext-SHAP-LIME-Urdu

The rendering layer modifies plotting code from the SHAP and LIME distributions, both released under the MIT License, and the derivative is released under the same terms with attribution to the original projects. The fonts distributed with the package are licensed under the SIL Open Font License. The four benchmark datasets used in this study are the property of their original authors and are available under the terms specified in [18, 25, 26].